\pdfoutput=1

\documentclass[10pt,twocolumn,letterpaper]{article}

\usepackage{cvpr}
\usepackage{times}
\usepackage{epsfig}
\usepackage{graphicx}
\usepackage{amsmath}
\usepackage{amssymb}

\usepackage[breaklinks=true,bookmarks=false]{hyperref}

\cvprfinalcopy % *** Uncomment this line for the final submission

\def\cvprPaperID{****} % *** Enter the CVPR Paper ID here

\usepackage{algorithm}% The package I import.
\usepackage{algpseudocode}% The package I import.
\usepackage{graphicx}% The package I import.
\usepackage{booktabs}% The package I import.
\usepackage{subfig}% The package I import.

\usepackage{color}% The package I import.
\usepackage{xcolor} % The package I import {red/blue/green/black/white/cyan/magenta/yellow}
\usepackage{multirow} % The package I import

\definecolor{PROTOSS PYLON}{RGB}{0, 168, 255} % The package I import.
\definecolor{VANADYL BLUE}{RGB}{0, 151, 230} % The package I import.

\definecolor{PERIWINKLE}{RGB}{156, 136, 255} % The package I import.
\definecolor{MATT PURPLE}{RGB}{140, 122, 230} % The package I import.

\definecolor{RISE-N-SHINE}{RGB}{251, 197, 49} % The package I import.
\definecolor{NANOHANACHA GOLD}{RGB}{225, 177, 44} % The package I import.

\definecolor{DOWNLOAD PROGRESS}{RGB}{76, 209, 55} % The package I import.
\definecolor{SKIRRET GREEN}{RGB}{68, 189, 50} % The package I import.

\definecolor{SEABROOK}{RGB}{72, 126, 176} % The package I import.
\definecolor{NAVAL}{RGB}{64, 115, 158} % The package I import.

\definecolor{NASTURCIAN FLOWER}{RGB}{232, 65, 24} % The package I import.
\definecolor{HARLEY DAVIDSON ORANGE}{RGB}{194, 54, 22} % The package I import.

\definecolor{LYNX WHITE}{RGB}{245, 246, 250} % The package I import.
\definecolor{HINT OF PENSIVE}{RGB}{220, 221, 225} % The package I import.

\definecolor{BLUEBERRY SODA}{RGB}{127, 143, 166} % The package I import.
\definecolor{CHAIN GANG GREY}{RGB}{113, 128, 147} % The package I import.

\definecolor{MAZARINE BLUE}{RGB}{39, 60, 117} % The package I import.
\definecolor{PICO VOID}{RGB}{25, 42, 86} % The package I import.

\definecolor{BLUE NIGHTS}{RGB}{53, 59, 72} % The package I import.
\definecolor{ELECTROMAGNETIC}{RGB}{47,54,64} % The package I import.

\makeatletter
  \newcommand\figcaption{\def\@captype{figure}\caption}
  \newcommand\algcaption{\def\@captype{algorithm}\caption}
\makeatother

\usepackage{rotating} % The package I import.

\ifcvprfinal\fi
\begin{document}

%%%%%%%%% TITLE
\title{Task Augmentation by Rotating for Meta-Learning}

\author{%
  Jialin~Liu \\
  Cognitive Science Department\\
  Xiamen University\\
  Fujian, P. R. China 361005 \\
  \texttt{jialin@stu.xmu.edu.cn} \\
  % examples of more authors
   \and
   Fei Chao \\
   Cognitive Science Department \\
      Xiamen University \\
   Fujian, P. R. China 361005 \\
   \texttt{fchao@xmu.edu.cn} \\
   \and
   Chih-Min Lin\\
    Department of Electrical Engineering\\
   Yuan ze University \\
   Chung-Li, Tao-Yuan 320, Taiwan\\
      \texttt{cml@saturn.yzu.edu.tw} \\
}

\maketitle
%\thispagestyle{empty}

%%%%%%%%% ABSTRACT
\begin{abstract}
  Data augmentation is one of the most effective approaches for improving the accuracy of modern machine learning models, and it is also indispensable to train a deep model for meta-learning. In this paper, we introduce a task augmentation method by rotating, which increases the number of classes by rotating the original images 90, 180 and 270 degrees, different from traditional augmentation methods which increase the number of images.
  %{\color{orange}However, as one data augmentation approach, rotation has not been applied for meta-learning. This work we proposed the idea of a new kind data augmentation, we called it as task augmentation (referred to Task Aug), and show that rotation for task augmentation is better than rotation for traditional data augmentation in meta-learning.}
  %The basic idea of Task Aug is to increase the number of image classes rather than the number of images in each class. In contrast,
  With a larger amount of classes, we can sample more diverse task instances during training. Therefore, task augmentation by rotating allows us to train a deep network by meta-learning methods with little over-fitting. Experimental results show that our approach is better than the rotation for increasing the number of images and achieves state-of-the-art performance on miniImageNet, CIFAR-FS, and FC100 few-shot learning benchmarks. The code is available on \url{www.github.com/AceChuse/TaskLevelAug}.
\end{abstract}

%%%%%%%%% BODY TEXT
\section{Introduction}

Although the machine learning systems have achieved a human-level ability in many fields with a large amount of data, learning from a few examples is still a challenge for modern machine learning techniques. Recently, the machine learning community has paid significant attention to this problem, where few-shot learning is the common task for meta-learning (e.g., \cite{ravi2017optimization,finn2017model,vinyals2016matching,snell2017prototypical}). The purpose of few-shot learning is to learn to maximize generalization accuracy across different tasks with few training examples. %After training, the meta learner is able to control the training process of the based learner and make it able to learn to classify data or fit curve based on a few training examples without over-fitting.
In a classification application of the few-shot learning, tasks are generated by sampling from a conventional classification dataset; then, training samples are randomly selected from several classes in the classification dataset. In addition, a part of the examples is used as training examples and testing examples. Thus, a tiny learning task is formed by these examples. The meta-learning methods are applied to control the learning process of a base learner, so as to correctly classify on testing examples. %and the classes used during training and evaluation do not overlap in general.

Data augmentation is widely used to improve the training of deep learning models. Usually, data augmentation is regarded as an explicit form of regularization \cite{he2016deep,simonyan2014very,krizhevsky2012imagenet}. Data augmentation aims at artificially generating the training data by using various translations on existing data, such as: adding noises, cropping, flipping, rotation, translation, etc. The general idea of data augmentations is increasing the number of images by change data slightly to be different from original data, but the data still can be recognized by human. The new images involved in the classes are identical to the original data, we call this as Image Aug.

\begin{figure*}[t]
\begin{center}
\includegraphics[width=140mm]{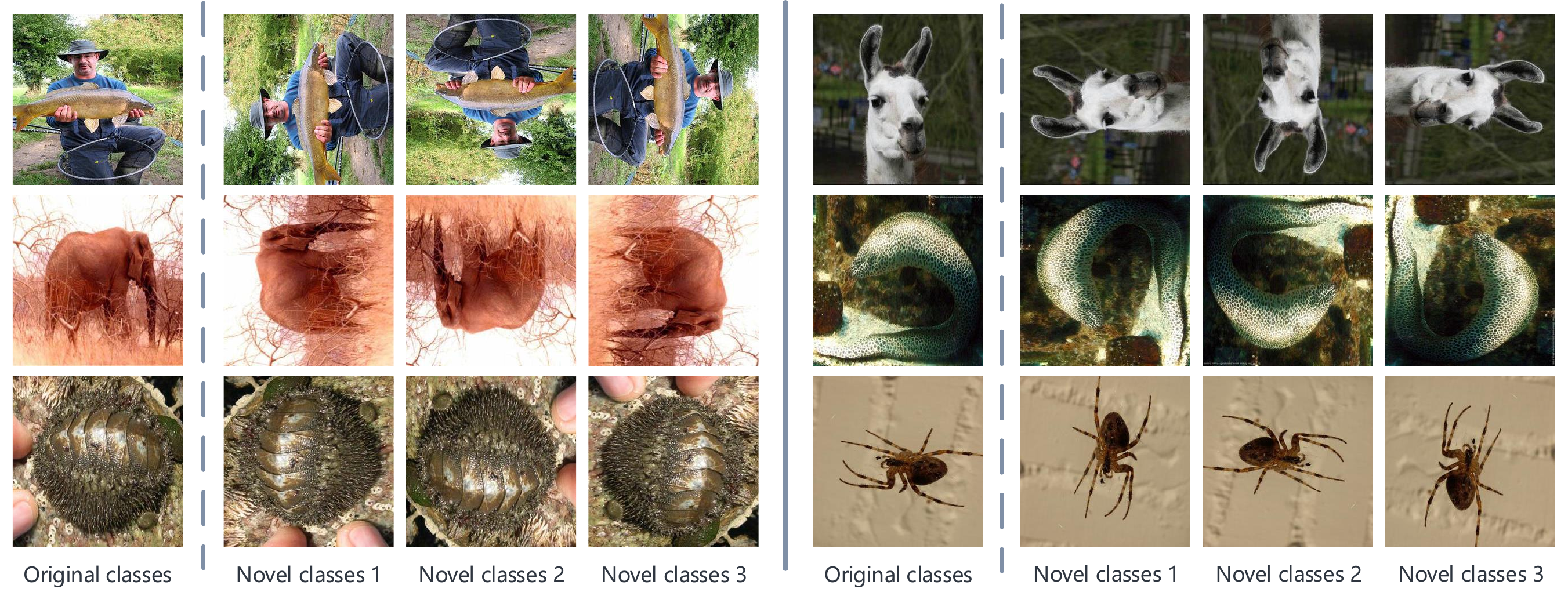}
\end{center}
\caption{Examples of the novel created classes.}
\label{TLAExamples}
\end{figure*}

However, the minimum units of meta-learning are tasks rather than data, so we should use rotation operation to augment the number of tasks, which is called as task augmentation (referred to Task Aug). Task Aug means increasing the types of task instances by increasing the data that can be clearly recognized as the different classes as the original data and associating them as the novel classes(we show examples in Figure~\ref{TLAExamples}). This is important for the meta-learning, since meta-learning models require to predict unseen classes during the testing phase, increasing the diverseness of tasks would help models to generate to unseen classes.

In experiments, we compared two cases, 1) the new images are converted to the classes of original images and 2) the new images are associated to the novel classes with the method proposed in \cite{bertinetto2018meta} on CIFAR-FS, FC100, miniImageNet few-shot learning tasks, and showed the second case got better results. Then the proposed method is evaluated by experiments with the state of art meta-learning methods~\cite{snell2017prototypical,lee2019meta,bertinetto2018meta} on CIFAR-FS, FC100, miniImageNet few-shot learning tasks, and compare with the results without the data augmentation by rotating. In the comparative experiments, Task Aug by rotating achieves the better accuracy than the original meta-learning methods. Besides, the best results of our experiments exceed the current state-of-art result over a large margin.

%------------------------------------------------------------------------
\section{Related Work}\label{Related_Work}
Meta-learning involves two hierarchies learning processes: low-level and high-level. The low-level learning process learns to deal with general tasks, often termed as the ``inner loop''; and the high-level learning process learns to improve the performance of a low-level task, often termed as the ``outer loop''. Since models are required to handle sensory data like images, deep learning methods are often applied for the ``outer loop''. However, the machine learning methods applied for the ``inner loop'' are very diverse. Based on different methods in the ``inner loop'', meta-learning can be applied in image recognition~\cite{fei2006one,santoro2016meta,finn2017model,vinyals2016matching,ravi2017optimization}, image generation~\cite{antoniou2017data,zhang2018metagan,rezende2016one}, reinforce learning~\cite{finn2017model,al2017continuous}, and etc. This work focuses on few-shot learning image recognition based on meta-learning. Therefore, in the experiment, the methods applied in the ``inner loop'' are able to classify data, and they are K-nearest neighbor (KNN), Support Vector Machine (SVM) and ridge regression, respectively \cite{snell2017prototypical,lee2019meta,bertinetto2018meta}.

Previous studies have introduced many popular regularization techniques to few-shot learning from deep learning, such as weight decay, dropout, label smooth~\cite{bertinetto2018meta}, and data augmentation. Common data augmentation techniques for image recognition are usually designed manually and the best augmentation strategies depend on dataset. In natural color image datasets, random cropping and random horizontal flipping are the most common. Since the few-shot learning tasks consist of natural color images, the random horizontal flipping and random cropping are applied in few-shot learning. In addition, color (brightness, contrast, and saturation) jitter is often applied in the works of few-shot learning~\cite{gidaris2018dynamic,qiao2018few}.

Other data augmentation technologies related to few-shot learning include generating samples by few-shot learning and generating samples for few-shot learning. The former tried to synthesize additional examples via transferring, extracting, and encoding to create the data of the new classes, that are intra-class relationships between pairs of reference classes' data instances~\cite{hariharan2017low,schwartz2018delta}. The later tried to apply meta-learning in a few-shot generation to generate samples from other models~\cite{antoniou2017data}.In addition to these two types of studies, the data augmentation technology most closed to the new proposed approach is applied to Omniglot dataset, which consists of handwritten words \cite{lake2015human}. They created the novel classes by rotating the original images 90, 180 and 270 degrees~\cite{santoro2016meta}. However, when this approach is applied  for the natural color image, it would be slightly changed, and we will explain this in Section~\ref{Method}.

%------------------------------------------------------------------------
\section{Method}\label{Method}

%\begin{algorithm}[t]
%  \caption{Task Level Data Augmentation.}\label{Ari_Task_Aug}
%  \begin{algorithmic}[1]
%    \Require
%      Classes set $\mathcal{C}=\{c_1,c_2,\dots,c_{M}\}$;
%      Max possibility for selecting the novel classes $p_{max}$;
%      The number of task instances during increasing probability $\rm T$; The number of currently generated task instances $t$; The number of ways, shots and queries $N$, $K$, $H$
%      \State $t\gets t+1$
%      \State $p\gets p_{max}*\min\{1,\frac{t}{\rm T}\}$
%      \State $n\sim Binomial(N,p)$
%      \State $\mathcal{D}^{tr},\mathcal{D}^{val}\gets\{\},\{\}$
%      \State $\mathcal{V}\gets$ Sample $N-n$ from $\{1,2,\cdots,M\}$
%      \ForAll {$v\in\mathcal{V}$}
%        \State $\mathcal{D}\gets$ Sample $K+H$ from $c_v$
%        \State $\mathcal{D}^{tr}\gets\mathcal{D}^{tr}\cup$ First $K$ of $\mathcal{D}$
%        \State $\mathcal{D}^{val}\gets\mathcal{D}^{val}\cup$ Last $H$ of $\mathcal{D}$
%      \EndFor
%      \State $\mathcal{U}\gets$ Sample $n$ from $\{M,M+1,\cdots,4M\}$
%      \ForAll {$u\in\mathcal{U}$}
%        \State $v\gets (u\mod M)+1$
%        \State $\mathcal{D}\gets$ Sample $K+H$ from $c_v$
%        \State $r\gets \lfloor\frac{u}{M}\rfloor$
%        \State Rotate all ${\rm x}\in\{{\rm x}|({\rm x},{\rm y})\in\mathcal{D}\}$ $90r$ degrees
%        \State $\mathcal{D}^{tr}\gets\mathcal{D}^{tr}\cup$ First $K$ of $\mathcal{D}$
%        \State $\mathcal{D}^{val}\gets\mathcal{D}^{val}\cup$ Last $H$ of $\mathcal{D}$
%      \EndFor
%      \State \Return $(\mathcal{D}^{tr},\mathcal{D}^{val})$
%  \end{algorithmic}
%\end{algorithm}

\begin{figure*}
\begin{minipage}{.52\textwidth}
  \begin{algorithm}[H] % Here must be [H]
  \algcaption{Task Augmentation by Rotating.}\label{Ari_Task_Aug}
  \begin{algorithmic}[1]
    \Require
      Classes set $\mathcal{C}=\{c_1,c_2,\dots,c_{M}\}$;
      Max possibility for Task Aug $p_{max}$;
      The delay to Task Aug $\rm T$; The current count $t$; The number of ways, shots and queries $N$, $K$, $H$
      \State $t\gets t+1$
      \State $p\gets p_{max}*\min\{1,\frac{t}{\rm T}\}$
      \State $n\sim Binomial(N,p)$
      \State $\mathcal{D}^{tr},\mathcal{D}^{val}\gets\{\},\{\}$
      \State $\mathcal{V}\gets$ Sample $N-n$ from $\{1,2,\cdots,M\}$
      \ForAll {$v\in\mathcal{V}$}
        \State $\mathcal{D}\gets$ Sample $K+H$ from $c_v$
        \State $\mathcal{D}^{tr}\gets\mathcal{D}^{tr}\cup$ First $K$ of $\mathcal{D}$
        \State $\mathcal{D}^{val}\gets\mathcal{D}^{val}\cup$ Last $H$ of $\mathcal{D}$
      \EndFor
      \State $\mathcal{U}\gets$ Sample $n$ from $\{M,M+1,\cdots,4M\}$
      \ForAll {$u\in\mathcal{U}$}
        \State $v\gets (u\mod M)+1$
        \State $\mathcal{D}\gets$ Sample $K+H$ from $c_v$
        \State $r\gets \lfloor\frac{u}{M}\rfloor$
        \State Rotate all ${\rm x}\in\{{\rm x}|({\rm x},{\rm y})\in\mathcal{D}\}$ $90r$ degrees
        \State $\mathcal{D}^{tr}\gets\mathcal{D}^{tr}\cup$ First $K$ of $\mathcal{D}$
        \State $\mathcal{D}^{val}\gets\mathcal{D}^{val}\cup$ Last $H$ of $\mathcal{D}$
      \EndFor
      \State \Return $(\mathcal{D}^{tr},\mathcal{D}^{val})$
  \end{algorithmic}
  \end{algorithm}
\end{minipage}
\begin{minipage}{.45\textwidth}
\begin{center}
  \includegraphics[width=60mm]{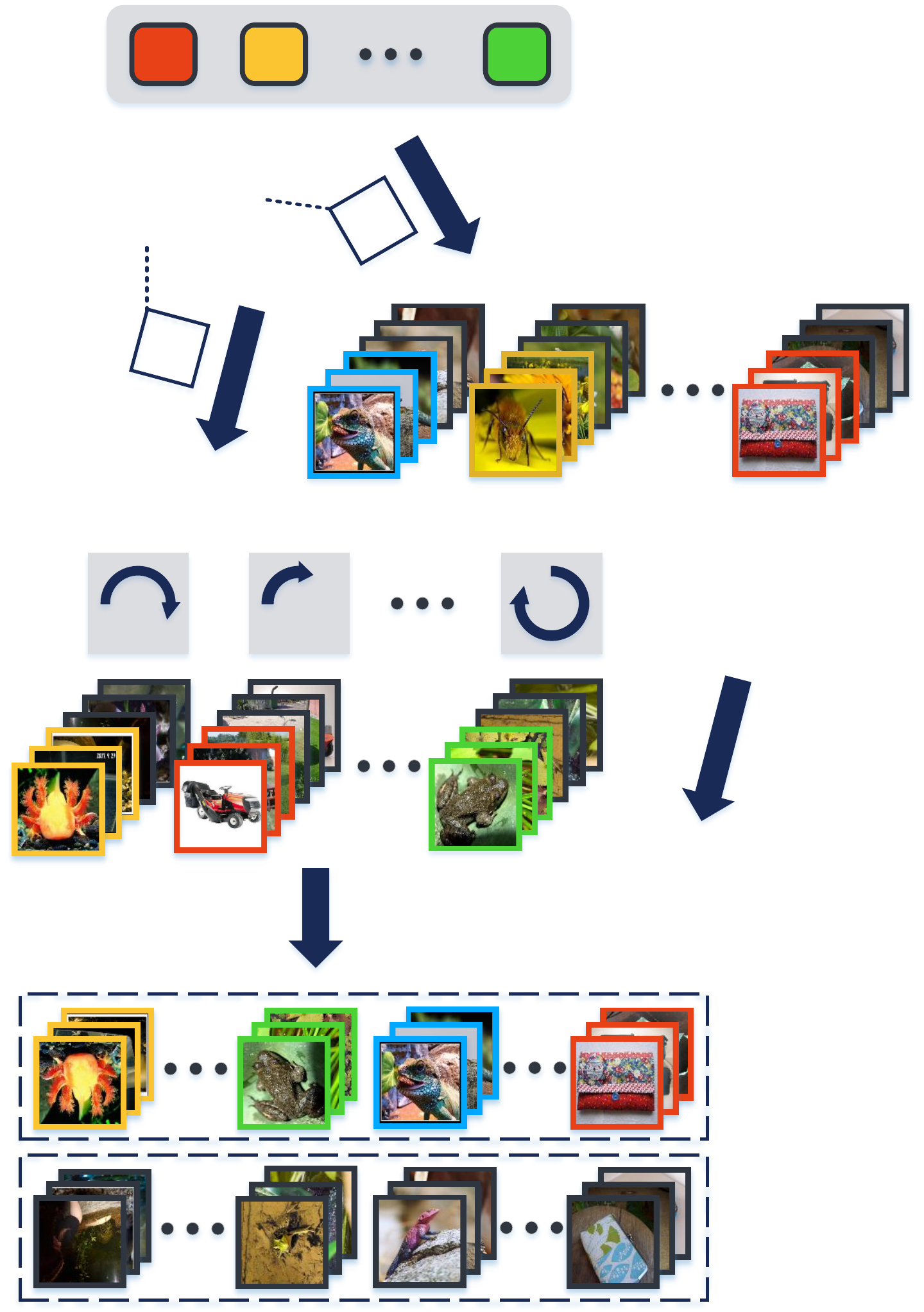}
  \put(-144,231){\small\color{ELECTROMAGNETIC}$c_1$}
  \put(-124,231){\small\color{ELECTROMAGNETIC}$c_2$}
  \put(-80,231){\small\color{ELECTROMAGNETIC}$c_M$}
  \put(-142,177){\begin{rotate}{-8}\small\color{PICO VOID}$p$\end{rotate}}
  \put(-106,196){\begin{rotate}{34}\footnotesize\color{PICO VOID}1-$p$\end{rotate}}
  \put(-162,210){\footnotesize\color{PICO VOID}$p_{max}*$}
  \put(-162,201){\footnotesize\color{PICO VOID}$\min\{1,\frac{t}{\rm T}\}$}
  \put(-75,191){\small\color{PICO VOID}$N$-$n$ from original classes}
  \put(-170,143){\small\color{PICO VOID}$n$ from novel classes}
  \put(-32,43){\color{PICO VOID}$\mathcal{D}^{tr}$}
  \put(-32,13){\color{PICO VOID}$\mathcal{D}^{val}$}
\end{center}
\figcaption{The process of generating a task instance with Task Aug by rotating.}
\label{TLA}
\end{minipage}
\end{figure*}
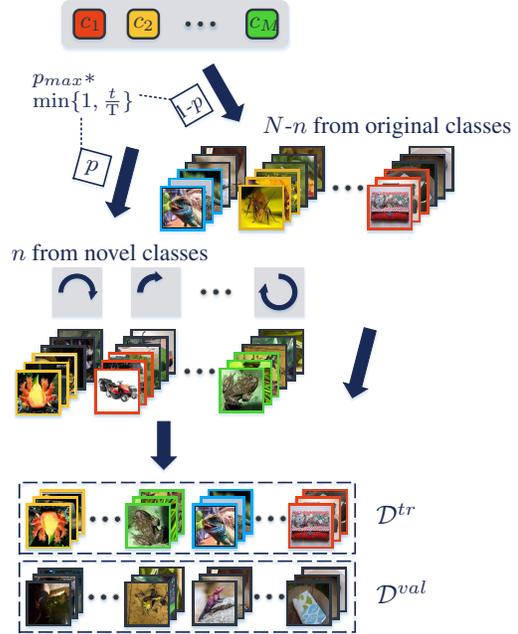

\subsection{Problem Definition}
We adopt the formulation purposed by \cite{vinyals2016matching} to describe the $N$-way $K$-shot task. A few-shot task contains many task instances (denoted by $\mathcal{T}_i$), each instance is a classification problem consisting of the data sampled from $N$ classes. The classes are randomly selected from a classes set. The classes set are split into $M^{tr}$, $M^{val}$ and $M^{test}$ for a training class set $\mathcal{C}^{tr}$, a validation classes set $\mathcal{C}^{val}$, and a test classes set $\mathcal{C}^{test}$. In particular, each class does not overlap others (i.e., the classes used during testing are unseen classes during training). Data is randomly sampled from $\mathcal{C}^{tr}$, $\mathcal{C}^{val}$ and $\mathcal{C}^{test}$, so as to create task instances for training meta-set $S^{tr}$, validation meta-set $S^{val}$, and test meta-set $S^{test}$, respectively. The validation and testing meta-sets are used for model selection and final evaluation, respectively.

The data in each task instance, $\mathcal{T}_i$, are divided into training examples $D^{tr}$ and validation examples $D^{val}$. Both of them only contains the data from $N$ classes which sampled from the appropriate classes set randomly (for a task instance applied during training, the classes form a subset of the training classes set $\mathcal{C}^{tr}$). In most settings, the training set $\mathcal{D}^{tr}=\{({\rm x}^k_n,{\rm y}^k_n)|n=1\dots N; k=1\dots K\}$ consists of $K$ data instances from each class, this processing usually called as a ``shot''. The validation set, $\mathcal{D}^{val}$, consists of several other data instances from the same classes, this processing is usually called as a ``query''. An evaluation is provided for generalization performance on the $N$ classification task instance $\mathcal{D}^{tr}$. Note that: the validation set of a task instance $\mathcal{D}^{val}$ (for optimizing model during ``outer loop'') is different from the held-out validation classes set $\mathcal{C}^{val}$ and meta-set $\mathcal{S}^{val}$ (for model selection).

\subsection{Task Augmentation by Rotating}
This work is to increase the size of the training classes set, $M^{tr}$, by rotating all images within the training classes set with 90, 180, 270 degrees. The size, $M^{tr}$, is increased for three times. In the Omniglot dataset consisting of handwritten words~\cite{santoro2016meta}, this approach works well, since it can rotate a handwritten word multiple of 90 degrees and treat the new one as another word; in addition, it is really possible that the novel word is similar to some words, which are not included in the training classes but existed. %However, for natural images, it is not the same cause. For examples, the images in the third line of Figure~\ref{TLAExamples} are difficult to identify which images are rotated. Moreover, the images are rarely rotated in the photos taken by humans.

For natural images, it is obvious that the images generated by rotating is real enough. But should the new generated images be classified as the novel classes or the original classes? It dependents on the similarity between the new images and the original classes. If the most of they are similar enough, the new images should be classified as the original classes, and vice versa. This logic shows that one of the two methods must be effective. Since there are almost not works merge the new images into the original classes which worked well, we assume that Task Aug by rotating is effective for meta-learning, and we will compare two methods in experiments.

%{\color{orange}There are two ways to augment data by rotating, increasing size of training classes set and increasing the number of images. In the Omniglot dataset consisting of handwritten words~\cite{santoro2016meta}, rotation for Task Aug works well, since it can rotate a handwritten word multiple of 90 degrees and treat the new one as another word; in addition, it is really possible that the novel words are similar to some words, which are not included in the training classes but existed.}

%{\color{orange}So the first problem should be considered, for natural images, rotation for Task Aug and Data Aug, which one is better? It is obvious that the images generated by rotating is real enough. But should the new generated images be classified as the novel classes or the original classes? It dependents on the similarity between the new images and the original classes. If the most of they are similar enough, the new images should be classified as the original classes, and vice versa. This logic shows that one of the two methods must be effective. Since there are almost not works increasing the number of images by rotating for meta-learning, we assume that rotating for Task Aug is effective for meta-learning, and we will compare two methods in experiments.}

Besides, it is different from in handwritten that we assign the new data smaller weights than the original data, so as to make models prioritize learning the features of the original classes, since the images generated by rotating rarely exist in the original data. This way makes the features of the novel classes as a supplement to prevent the augmented data from taking up large capacity in the model, which is same as other common data augment methods.

%Despite the two problems, the fundamental features of the novel images can provide useful information. We assign novel classes that contain smaller weights than the original classes, so as to make models prioritize learning the features of the original classes, and make the features of the novel classes as a supplement to prevent the augmented data from taking up large capacity in the model. %{\color{orange}and the two problems can be solved by assigning the novel classes smaller weights than the original classes. When the images sampled for different classes are similar, the best strategy is classifying the images into the class with the higher weight or higher frequency.}

The smaller weights are implemented in two ways, 1) lower probability and 2) delaying the probability of selecting the novel classes. For a class in a task instance, the probability of the class coming from the novel classes is $p$, and the probability coming from the original classes is $1-p$. Besides, the initial $p$ is set to $0$, then linearly rises from $0$ to $p_{max}$ for the first $\rm T$ tasks. The max probability $p_{max}$ is set lower than the proportion of the novel classes in all classes to make each novel class have a lower probability than each original class. The whole process of Task Aug on a classes set is summarized in Algorithm~\ref{Ari_Task_Aug} and Figure~\ref{TLA}.

\subsection{Ensemble}
In this work, we also compare the methods with the training protocol with ensemble method~\cite{huang2017snapshot} in addition to the standard training protocol, which choosing a model by the validation set. The training protocol with an ensemble method use the models with different training epoch to an ensemble model, in order to better use the models obtained in a single training process, and this approach has been proved to be valid for meta-learning by experiments~\cite{liu2018learning}. We adopt this ensemble method. However, unlike \cite{huang2017snapshot} and \cite{liu2018learning} that we did not use cyclic annealing for learning rate and any methods to select models. We directly took the average of the prediction of all models, which are saved according to an interval of 1 epoch. In Section~\ref{experiments}, the methods with this ensemble approach are marked by ``+ens''.

\begin{figure*}[t]
\begin{center}
\subfloat{
\includegraphics[width=56mm]{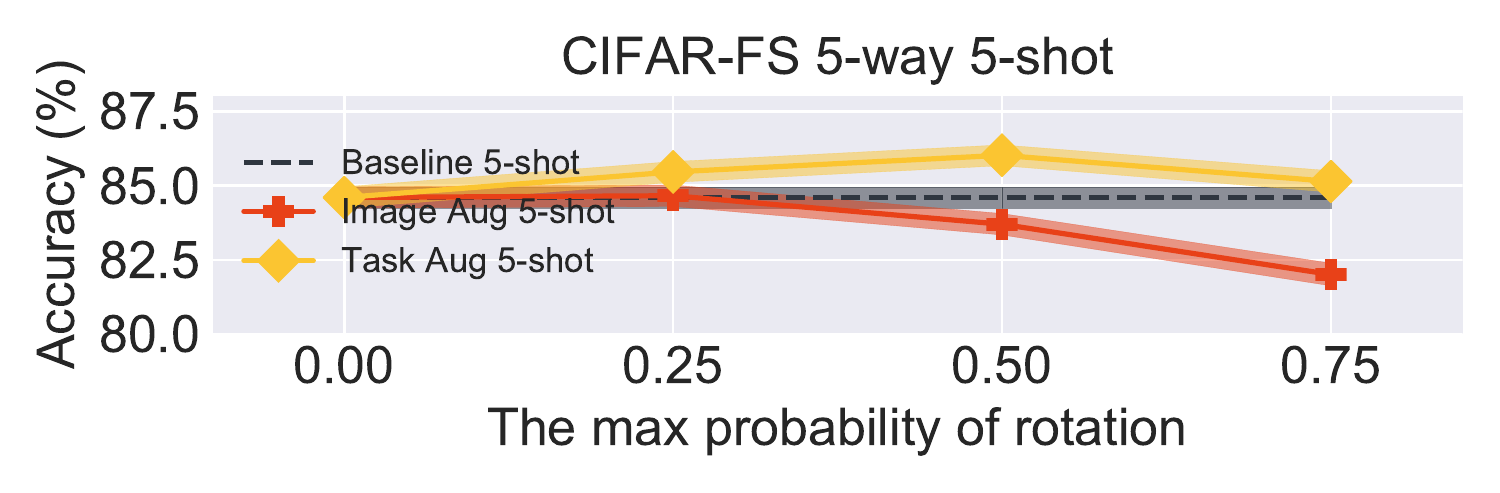}
\label{CIFAR-FS_y_5shot}}
\subfloat{
\includegraphics[width=56mm]{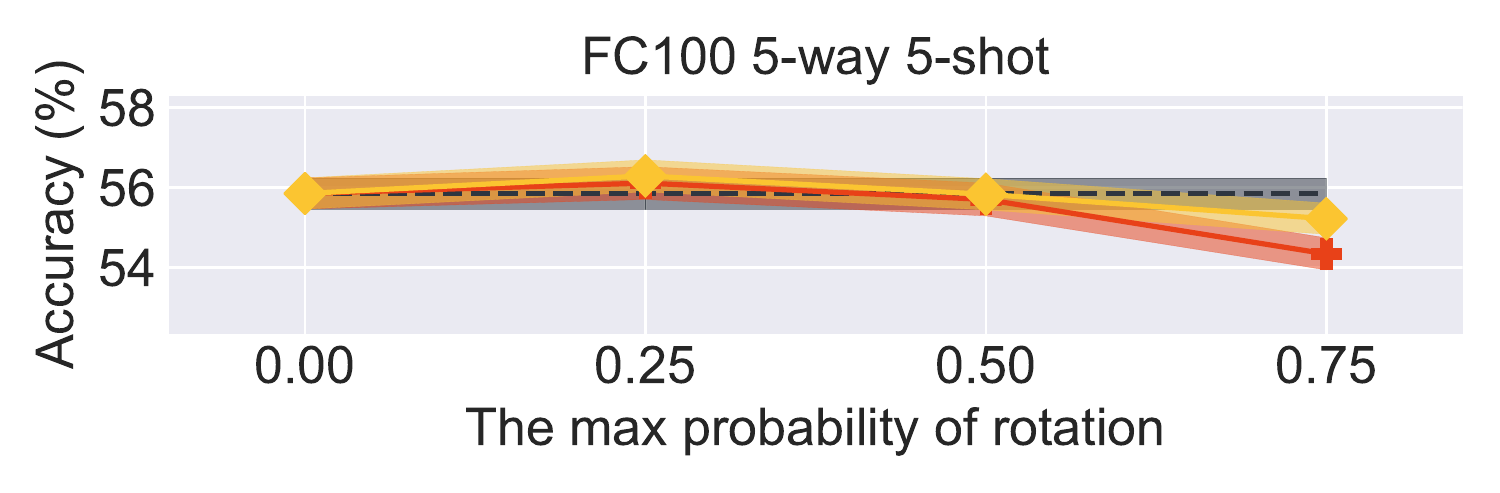}
\label{FC100_y_5shot}}
\subfloat{
\includegraphics[width=56mm]{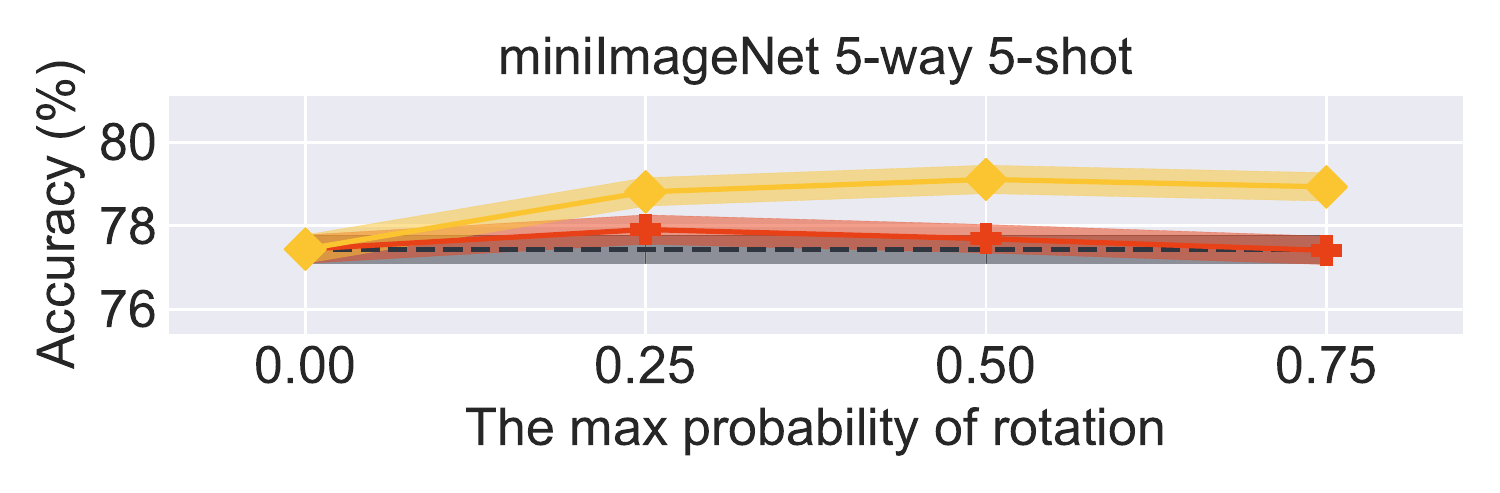}
\label{miniImageNet_y_5shot}}

\subfloat{
\includegraphics[width=56mm]{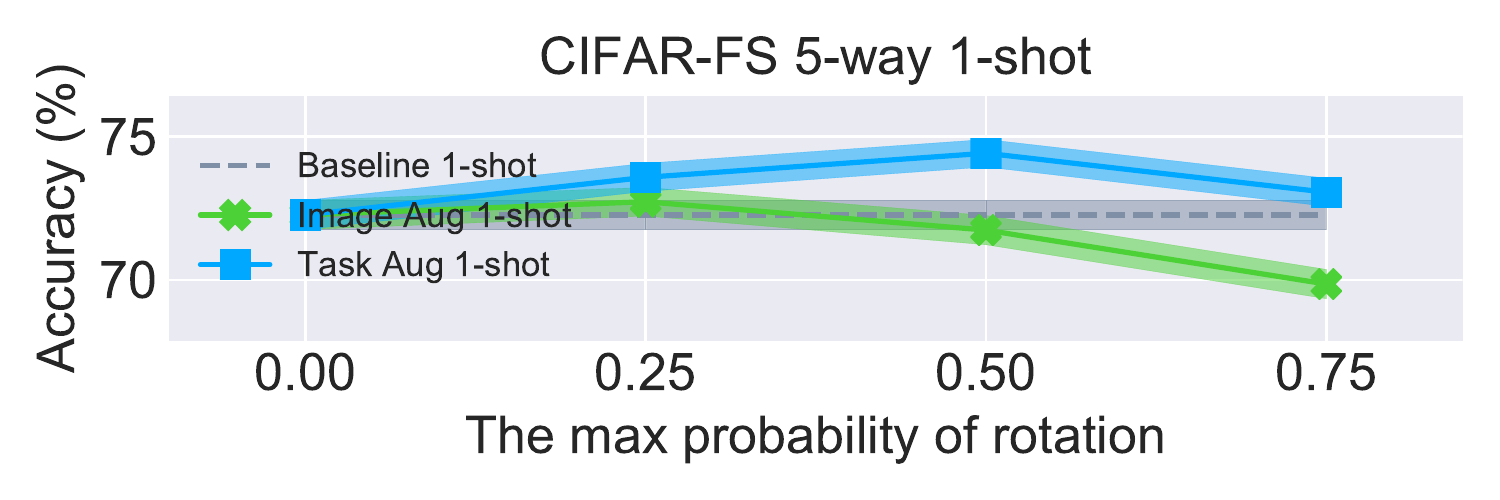}
\label{CIFAR-FS_y_1shot}}
\subfloat{
\includegraphics[width=56mm]{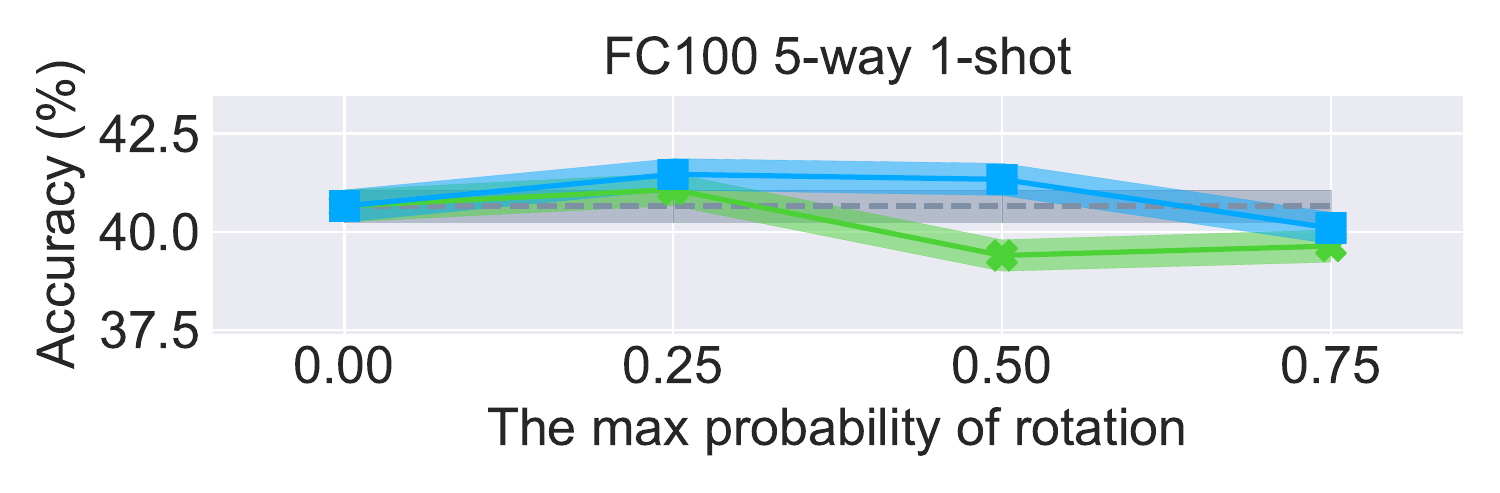}
\label{FC100_y_1shot}}
\subfloat{
\includegraphics[width=56mm]{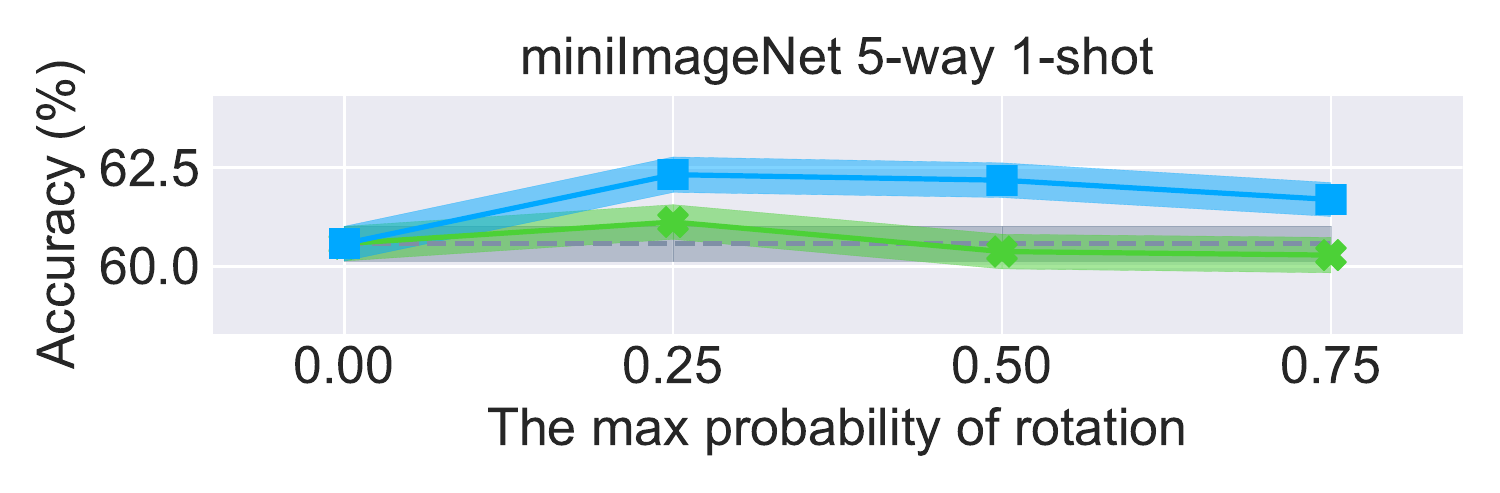}
\label{miniImageNet_y_1shot}}

\subfloat{
\includegraphics[width=56mm]{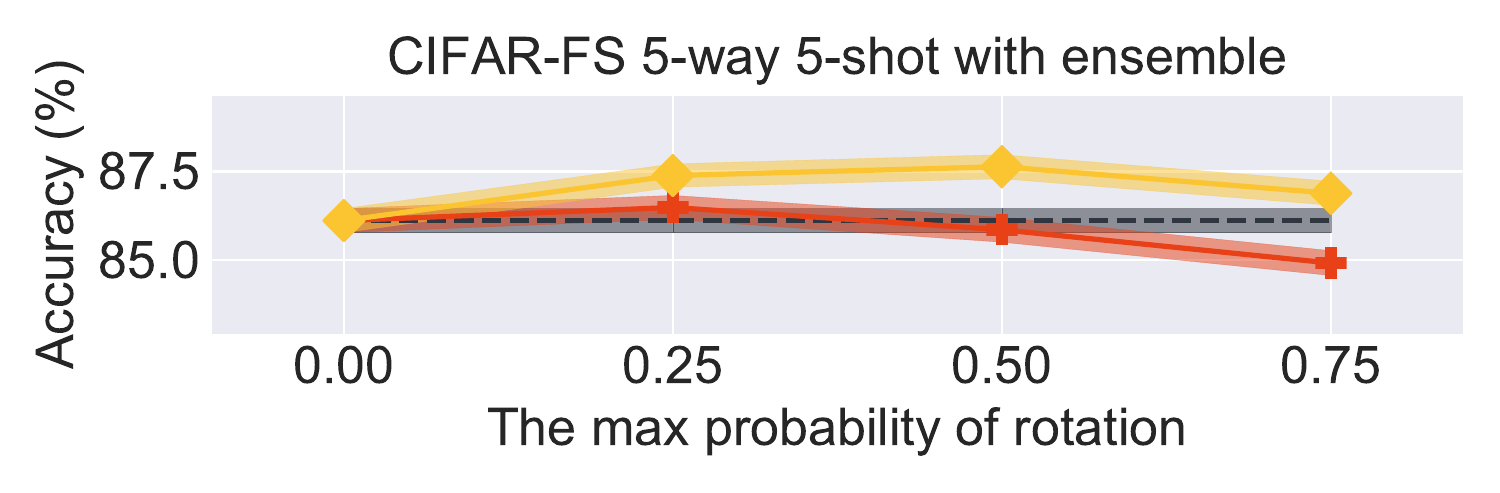}
\label{CIFAR-FS_ens_5shot}}
\subfloat{
\includegraphics[width=56mm]{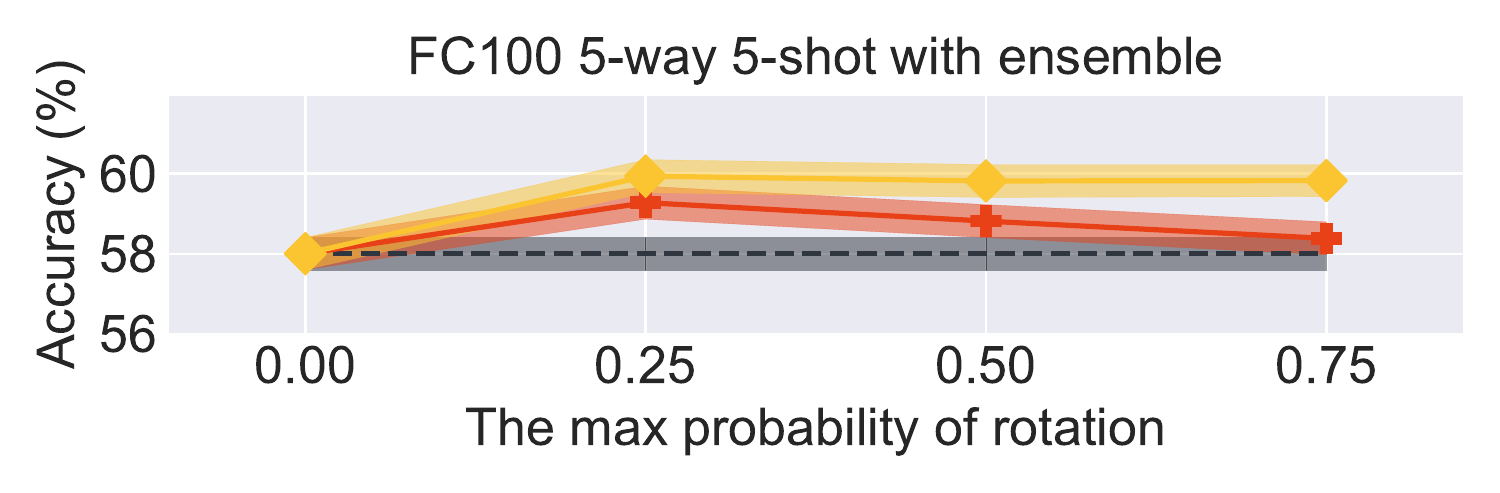}
\label{FC100_ens_5shot}}
\subfloat{
\includegraphics[width=56mm]{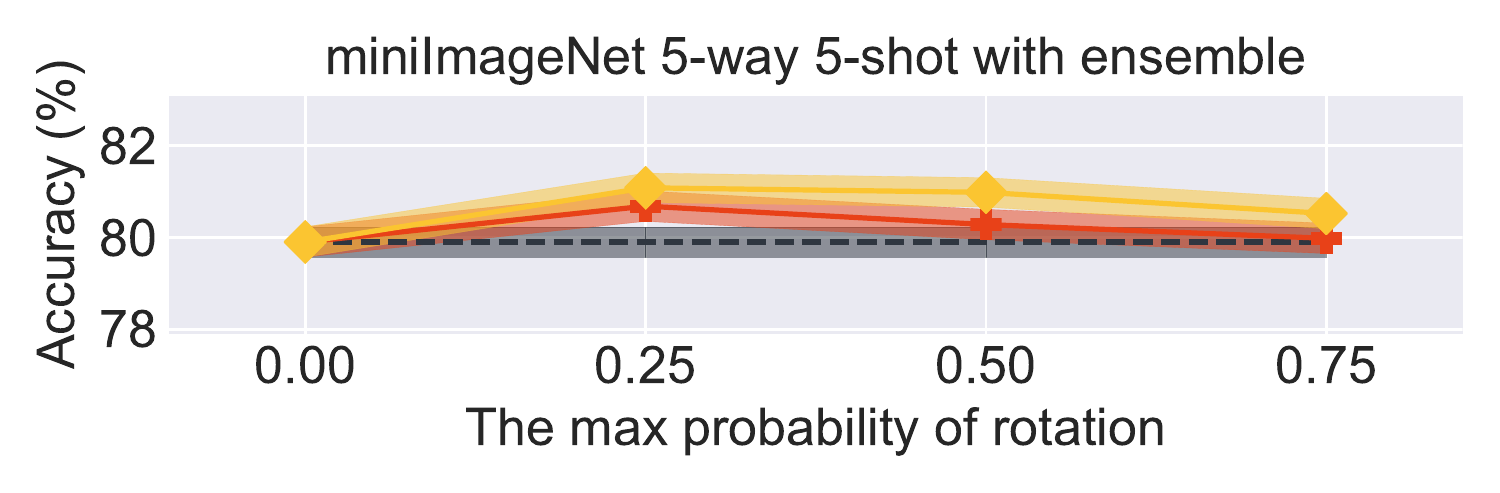}
\label{miniImageNet_ens_5shot}}

\subfloat{
\includegraphics[width=56mm]{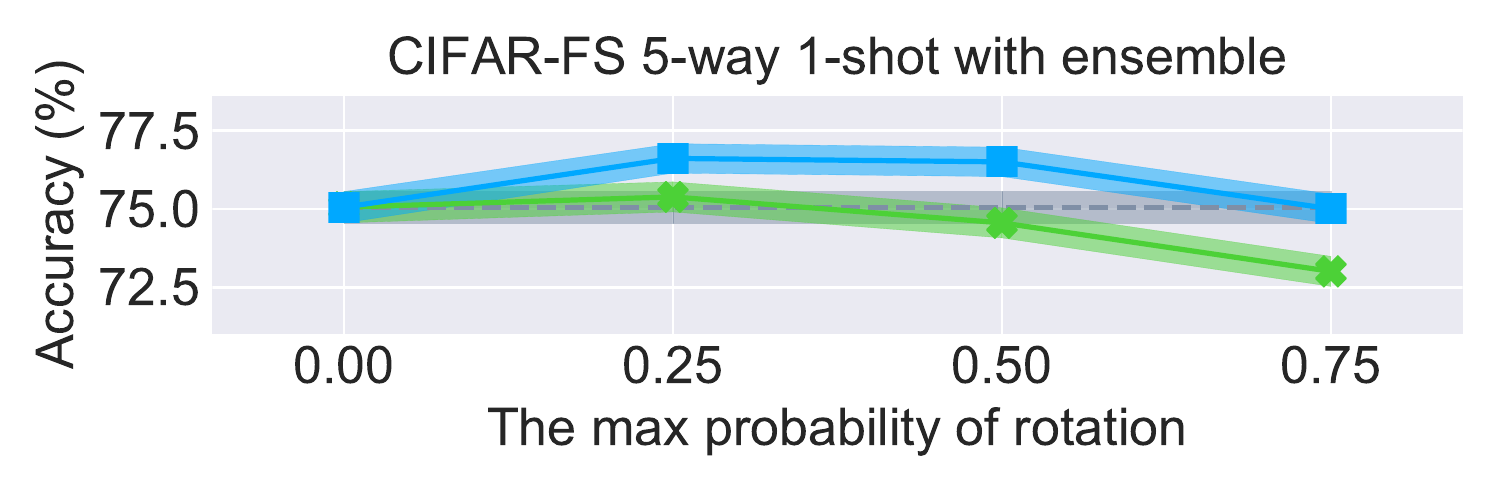}
\label{CIFAR-FS_ens_1shot}}
\subfloat{
\includegraphics[width=56mm]{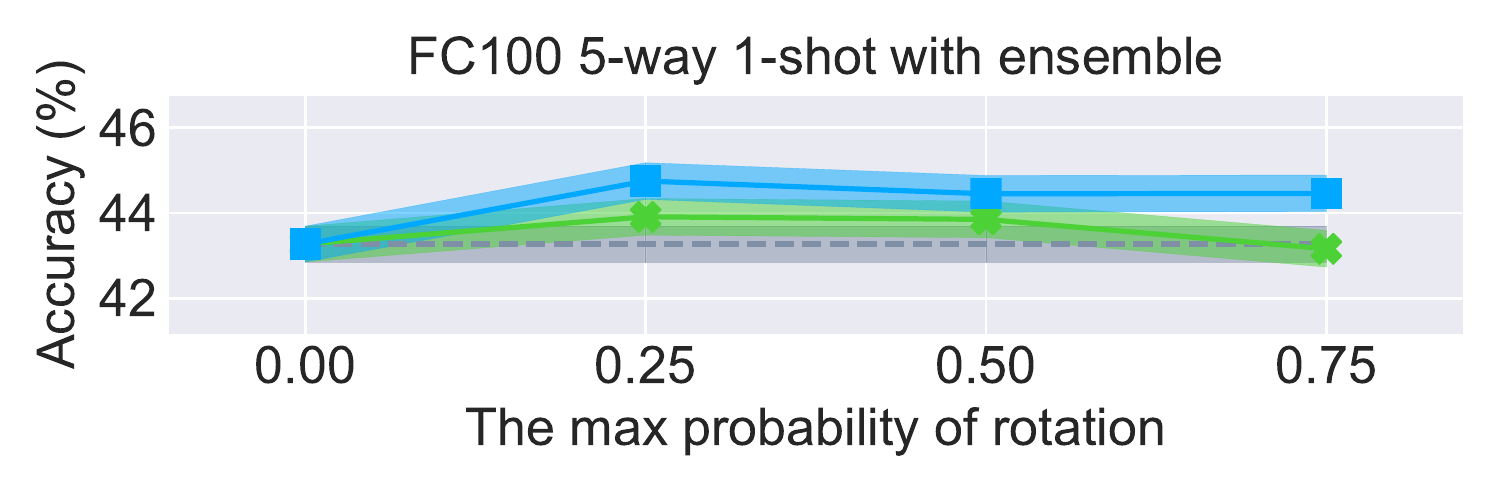}
\label{FC100_ens_1shot}}
\subfloat{
\includegraphics[width=56mm]{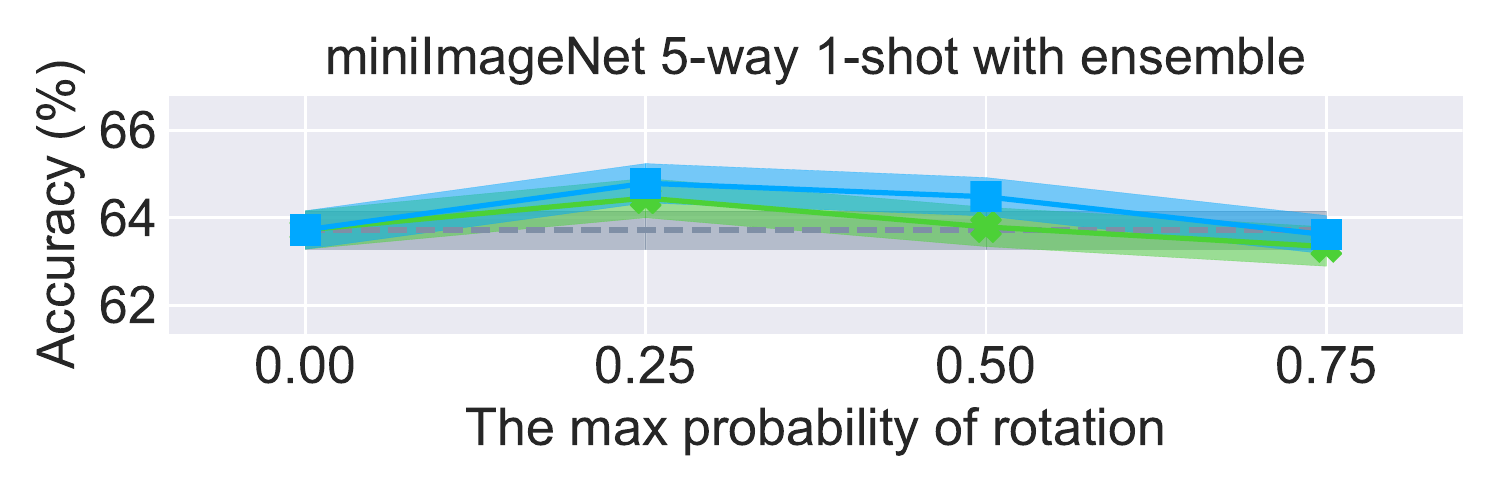}
\label{miniImageNet_ens_1shot}}
\end{center}
   \caption{The accuracies (\%) on meta-test sets with varying probability $p_{max}$ for the novel classes.The 95\% confidence interval is denoted by the shaded region.}
\label{Task_Aug_p}
\end{figure*}

\section{Experiments}\label{experiments}
We evaluate the proposed method on few-shot learning tasks. In order to ensure fair, both the results of baseline and Task Aug were run in our own environment. The comparative experiment is designed to answer the following questions: (1) Image Aug and Task Aug by rotating which is able to improve the performance of meta-learning? (2) How much should the probably for the novel classes be set? (3) Is Task Aug by rotating able to improve the performance of the current popular meta-learning methods?

\subsection{Experimental Configuration}
\subsubsection{Backbone}
Following \cite{lee2019meta,oreshkin2018tadam,mishra2017simple}, we used ResNet-12 network in our experiments. The ResNet-12 network had four residual blocks which contains three $3\times3$ convolution, batch normalization and Leaky ReLU with 0.1 negative slope. One $2\times2$ max-pooling layer is used for reducing the size of the feature map. The numbers of the network channels were 64, 160, 320 and 640, respectively. DropBlock regularization~\cite{ghiasi2018dropblock} is used in the last two residual blocks, the conventional dropout~\cite{hinton2012improving} is used in the first two residual blocks. The block sizes of DropBlock were set to 2 and 5 for CIFAR derivatives and ImageNet derivatives, respectively. In all experiments, the dropout possibility was set to 0.1. The global average pooling was not used for the final output of the last residual block.

\subsubsection{Base Learners}\label{Base_Learners}
We used ProtoNets~\cite{snell2017prototypical}, MetaOptNet-SVM~\cite{lee2019meta} (we write it as M-SVM) and Ridge Regression Differentiable Discriminator (R2-D2)~\cite{bertinetto2018meta} as basic methods to verify the effective of Task Aug.

For ProtoNets, we did not use a higher way for training than testing like \cite{snell2017prototypical}. Instead, the equal number of shot and way were used in both training and evaluation, and its output multiplied by a learnable scale before the softmax following \cite{oreshkin2018tadam,lee2019meta}.

For M-SVM, we set training shot to 5 for CIFAR-FS; 15 for FC100; and 15 for miniImageNet; %10 for tieredImageNet;
regularization parameter of SVM was set to 0.1; and a learnable scale was used following \cite{lee2019meta}. We did not use label smoothing like \cite{lee2019meta}, because we did not find that label smoothing can improve the performance in our environment. This was also affirmed from the \cite{lee2019meta} author's message on GitHub, that Program language packages and environment might affect results of the meta-learning method.

For R2-D2, we set the same training shot as for M-SVM, and used a learnable scale and bias following \cite{bertinetto2018meta}. It was different from \cite{bertinetto2018meta} we used a fixed regularization parameter of ridge regression which was set to 50 because \cite{bertinetto2018meta} has confirmed that making it learnable might not be helpful.

Last, for all methods, each class in a task instance contained 6 test (query) examples during training and 15 test (query) examples during testing.

\subsubsection{Training Configuration}
Stochastic gradient descent (SGD) was used. Following \cite{sutskever2013importance}, we set weight decay and Nesterov momentum to 0.0005 and 0.9, respectively. Each mini-batch contained 8 task instances. The meta-learning model was trained for 60 epochs, and 1000 mini-batchs for each epoch. We set the initial learning rate to 0.1, then multiplied it by 0.06, 0.012, and 0.0024 at epochs 20, 40 and 50, respectively, as in \cite{gidaris2018dynamic}. The results, which are marked by ``+ens'' were used the 60 models saved after each epoch to become an ensemble model. For the final training, the training classes set was augmented by the validation classes set. When we only chose one model, we will chose the model at the epoch where we got the best model during training on the training classes set. The results of the final run are marked by ``+val'' in this subsection. Since the base idea of ``+ens'' was proposed by other works and ``+val'' is popular for meta-learning, we do not explain more details about them.

For data augmentation, we adopted random crop, horizontal flip, and color (brightness, saturation, and contrast) jitter data augmentation following the work of \cite{gidaris2018dynamic,qiao2018few}. In the experiments of comparing Task Aug and Image Aug by rotating, R2-D2 was applied, and we set ${\rm T}$ to 80000. In the evaluation of Task Aug for ProtoNets and M-SVM, we set $p_{max}$ to the value getting the best results for R2-D2.
%We set $p_{max}$ to 0.5 for CIFAR-FS; 0.25 for FC100 and miniImageNet; and ${\rm T}$ was set to 80000 for all experiments.

\subsubsection{Dataset}
The \textbf{CIFAR-FS}~\cite{bertinetto2018meta} containing all 100 classes from CIFAR-100~\cite{krizhevsky2010cifar} is proposed as few-shot classification benchmark recently. These classes are randomly divided into training classes, validation classes and test classes. The three types contain 64, 16 and 20 classes, respectively. There are 600 nature color images of size $32\times32$ in each class.

The \textbf{FC100}~\cite{oreshkin2018tadam} are also derived from CIFAR-100~\cite{krizhevsky2010cifar}, and the 100 classes are grouped into 20 superclasses. The training, validation, and testing classes contain 60 classes from 12 superclasses, 20 classes from 4 superclasses, and 20 classes from 4 superclasses, respectively. The target is to minimize the information overlap between classes to make it more challenging than current few-shot classification tasks. Same as CIFAR-FS, there are 600 nature color images of size $32\times32$ in each class.

The \textbf{miniImageNet}~\cite{vinyals2016matching} is one of the most popular benchmark for few-shot classification, which contains 100 classes randomly selected from ILSVRC-2012~\cite{russakovsky2015imagenet}. The classes are randomly divided into training classes, validation classes and test classes, and them contain 64, 16 and 20 classes, respectively. There are 600 nature color images of size $84\times84$ in each class. Since \cite{vinyals2016matching} did not release the class splits, we use the more common split proposed by \cite{ravi2017optimization}.

%{\color{orange}The \textbf{tieredImageNet} benchmark~\cite{ren2018meta} is a larger subset of ILSVRC-2012~\cite{russakovsky2015imagenet}, composed of 608 classes grouped into 34 high-level categories. These are divided into 20 categories for meta-training, 6 categories for meta-validation, and 8 categories for meta-testing. This corresponds to 351, 97 and 160 classes for meta-training, meta-validation, and meta-testing respectively. This dataset aims to minimize the semantic similarity between the splits. All images are of size $84\times84$.}

\subsection{Comparison between Task Aug and Image Aug}
To prove our assumption that rotation multi 90 degrees for Task Aug is better than that for Image Aug, we draw the accuracy curves depending on $p_{max}$ for both Task Aug and Image Aug (curves showed in Figure~\ref{Task_Aug_p}). %To identify whether the rotation multi 90 degrees for Task Aug is better than that for {\color{orange}Image} Aug, we analyzed the experiment on CIFAR-FS, FC100 and miniImageNet.
The linear rising of $p$ was also used for Image Aug, and ${\rm T}=80000$ for both Task Aug and Image Aug. In all the results showed in Figure~\ref{Task_Aug_p}, the training classes set was not augmented by the validation classes set.

As shown in Figure~\ref{Task_Aug_p}, the performance of Task Aug on most of the regimes is better than Image Aug and baseline in general. Besides, we observed that: with the increase of $p_{max}$, the accuracy rises at first, reaches the peaks between 0.25 and 0.5, then declines and reaches baseline when $p_{max}=0.75$ at the end, which is the proportion of the novel classes in all classes. The accuracy of Task Aug on CIFAR-FS, FC100 and miniImagNet reach the peaks at 0.5, 0.25 and 0.25 respectively. At the same time, the rotation multi 90 degrees for Image Aug cannot improve or even cause worse performance. %Although delaying the probability of selecting the novel classes only raise accuracy slightly, Task Aug is effective in general.

\subsection{Evaluation of Task Aug}
In order to further prove the proposed approach can improve the performance of meta-learning, we evaluate Task Aug by rotating on several meta-learning methods in this section.

We choose several currently the state of art base learners for experiments, we detail in Section~\ref{Base_Learners}. Besides, the training protocol with ensemble method can get better results than the standard training protocol, we involve it in the experiments. We think this is important, because the proposed method can only be a contribution if it can further improve performance based on the best method available at present.

\begin{table}
\caption{Comparison to the average accuracies (\%) with 95\% confidence intervals between the methods with and without Task Aug on \textbf{CIFAR-FS 5-way 1-shot}.}
\label{CIFAR-FS_1shot}
\begin{center}
\begin{tabular}{lcc}
\toprule[1pt]
\textbf{Method} & Baseline & Task Aug \\
\hline
ProtoNets~\cite{snell2017prototypical} & 71.88$\pm$0.52 & \textbf{74.15$\pm$0.50}\\
ProtoNets (+ens) & 73.95$\pm$0.51 & \textbf{75.89$\pm$0.48}\\
ProtoNets (+val) & 73.20$\pm$0.51 & \textbf{75.10$\pm$0.49}\\
ProtoNets (+ens+val) & 76.05$\pm$0.49 & \textbf{77.28$\pm$0.47}\\
\hline
M-SVM~\cite{lee2019meta} & 71.52$\pm$0.51 & \textbf{72.95$\pm$0.48}\\
M-SVM (+ens) & 74.12$\pm$0.50 & \textbf{75.85$\pm$0.47}\\
M-SVM (+val) & 72.42$\pm$0.50 & \textbf{73.13$\pm$0.47}\\
M-SVM (+ens+val) & 75.91$\pm$0.48 & \textbf{76.75$\pm$0.46}\\
\hline
R2-D2~\cite{bertinetto2018meta} & 72.27$\pm$0.51 & \textbf{74.42$\pm$0.48}\\
R2-D2 (+ens) & 75.06$\pm$0.50 & \textbf{76.51$\pm$0.47}\\
R2-D2 (+val) & 73.52$\pm$0.50 & \textbf{76.02$\pm$0.47}\\
R2-D2 (+ens+val) & 76.40$\pm$0.49 & \textbf{77.66$\pm$0.46}\\
\bottomrule[1pt]
\end{tabular}
\end{center}
\end{table}

\begin{table}[t]
\caption{Comparison to the average accuracies (\%) with 95\% confidence intervals between the methods with and without Task Aug on \textbf{CIFAR-FS 5-way 5-shot}.}
\label{CIFAR-FS_5shot}
\begin{center}
\begin{tabular}{lcc}
\toprule[1pt]
\textbf{Method} & Baseline & Task Aug \\
\hline
ProtoNets~\cite{snell2017prototypical} & 84.14$\pm$0.36 & \textbf{85.37$\pm$0.35}\\
ProtoNets (+ens) & 85.72$\pm$0.35 & \textbf{87.33$\pm$0.33}\\
ProtoNets (+val) & 85.29$\pm$0.35 & \textbf{86.53$\pm$0.34}\\
ProtoNets (+ens+val) & 86.88$\pm$0.34 & \textbf{88.24$\pm$0.33}\\
\hline
M-SVM~\cite{lee2019meta} & 84.01$\pm$0.36 & \textbf{85.91$\pm$0.36}\\
M-SVM (+ens) & 85.85$\pm$0.34 & \textbf{87.73$\pm$0.33}\\
M-SVM (+val) & 84.94$\pm$0.36 & \textbf{86.94$\pm$0.34}\\
M-SVM (+ens+val) & 87.15$\pm$0.34 & \textbf{88.38$\pm$0.33}\\
\hline
R2-D2~\cite{bertinetto2018meta} & 84.60$\pm$0.36 & \textbf{86.02$\pm$0.35}\\
R2-D2 (+ens) & 86.11$\pm$0.34 & \textbf{87.63$\pm$0.34}\\
R2-D2 (+val) & 85.39$\pm$0.36 & \textbf{86.73$\pm$0.34}\\
R2-D2 (+ens+val) & 87.04$\pm$0.34 & \textbf{88.33$\pm$0.33}\\
\bottomrule[1pt]
\end{tabular}
\end{center}
\end{table}

\begin{table}[t]
\caption{Comparison to the average accuracies (\%) with 95\% confidence intervals between the methods with and without Task Aug on \textbf{FC100 5-way 1-shot}.}
\label{FC100_1shot}
\begin{center}
\begin{tabular}{lcc}
\toprule[1pt]
\textbf{Method} & Baseline & Task Aug \\
\hline
ProtoNets~\cite{snell2017prototypical} & 37.53$\pm$0.40 & \textbf{38.89$\pm$0.40
}\\
ProtoNets (+ens) & 40.04$\pm$0.41 & \textbf{42.00$\pm$0.43}\\
ProtoNets (+val) & 43.63$\pm$0.43 & \textbf{44.91$\pm$0.46}\\
ProtoNets (+ens+val) & 47.16$\pm$0.46 & \textbf{48.91$\pm$0.47}\\
\hline
M-SVM~\cite{lee2019meta} & 40.50$\pm$0.39 & \textbf{41.17$\pm$0.40}\\
M-SVM (+ens) & 43.24$\pm$0.42 & \textbf{44.38$\pm$0.42}\\
M-SVM (+val) & 46.72$\pm$0.45 & \textbf{47.39$\pm$0.44}\\
M-SVM (+ens+val) & 49.50$\pm$0.46 & \textbf{49.77$\pm$0.45}\\
\hline
R2-D2~\cite{bertinetto2018meta} & 40.66$\pm$0.41 & \textbf{41.47$\pm$0.40}\\
R2-D2 (+ens) & 43.27$\pm$0.42 & \textbf{44.75$\pm$0.43}\\
R2-D2 (+val) & 47.12$\pm$0.44 & \textbf{48.21$\pm$0.45}\\
R2-D2 (+ens+val) & 49.92$\pm$0.45 & \textbf{51.35$\pm$0.46}\\
\bottomrule[1pt]
\end{tabular}
\end{center}
\end{table}

\begin{table}[t]
\caption{Comparison to the average accuracies (\%) with 95\% confidence intervals between the methods with and without Task Aug on \textbf{FC100 5-way 5-shot}.}
\label{FC100_5shot}
\begin{center}
\begin{tabular}{lcc}
\toprule[1pt]
\textbf{Method} & Baseline & Task Aug \\
\hline
ProtoNets~\cite{snell2017prototypical} & 51.43$\pm$0.39 & \textbf{53.92$\pm$0.39}\\
ProtoNets (+ens) & 54.24$\pm$0.40 & \textbf{56.55$\pm$0.40}\\
ProtoNets (+val) & \textbf{61.16$\pm$0.42} & 60.86$\pm$0.41\\
ProtoNets (+ens+val) & 63.64$\pm$0.43 & \textbf{65.47$\pm$0.42}\\
\hline
M-SVM~\cite{lee2019meta} & 54.83$\pm$0.40 & \textbf{56.23$\pm$0.40}\\
M-SVM (+ens) & 58.49$\pm$0.41 & \textbf{60.14$\pm$0.41}\\
M-SVM (+val) & 62.99$\pm$0.42 & \textbf{63.64$\pm$0.42}\\
M-SVM (+ens+val) & 66.37$\pm$0.42 & \textbf{67.17$\pm$0.41}\\
\hline
R2-D2~\cite{bertinetto2018meta} & 55.85$\pm$0.39 & \textbf{56.29$\pm$0.40}\\
R2-D2 (+ens) & 58.01$\pm$0.40 & \textbf{59.94$\pm$0.41}\\
R2-D2 (+val) & 63.32$\pm$0.40 & \textbf{64.53$\pm$0.42}\\
R2-D2 (+ens+val) & 65.58$\pm$0.42 & \textbf{67.66$\pm$0.42}\\
\bottomrule[1pt]
\end{tabular}
\end{center}
\end{table}

\begin{table}[t]
\caption{Comparison to the average accuracies (\%) with 95\% confidence intervals between the methods with and without Task Aug on \textbf{miniImageNet 5-way 1-shot}.}
\label{miniImageNet_1shot}
\begin{center}
\begin{tabular}{lcc}
\toprule[1pt]
\textbf{Method} & Baseline & Task Aug\\
\hline
ProtoNets~\cite{snell2017prototypical} & 58.67$\pm$0.48 & \textbf{60.52$\pm$0.48}\\
ProtoNets (+ens) & 62.12$\pm$0.48 & \textbf{63.69$\pm$0.47}\\
ProtoNets (+val) & 60.13$\pm$0.48 & \textbf{62.22$\pm$0.49}\\
ProtoNets (+ens+val) & 63.84$\pm$0.48 & \textbf{65.04$\pm$0.48}\\
\hline
M-SVM~\cite{lee2019meta} & 60.02$\pm$0.45 & \textbf{62.12$\pm$0.44}\\
M-SVM (+ens) & 63.44$\pm$0.45 & \textbf{64.56$\pm$0.44}\\
M-SVM (+val) & 61.58$\pm$0.45 & \textbf{63.14$\pm$0.45}\\
M-SVM (+ens+val) & 64.74$\pm$0.45 & \textbf{65.38$\pm$0.45}\\
\hline
R2-D2~\cite{bertinetto2018meta} & 60.57$\pm$0.44 & \textbf{62.32$\pm$0.45}\\
R2-D2 (+ens) & 63.72$\pm$0.44 & \textbf{64.79$\pm$0.45}\\
R2-D2 (+val) & \textbf{62.82$\pm$0.45} &62.64$\pm$0.44\\
R2-D2 (+ens+val) & 65.50$\pm$0.45 & \textbf{65.95$\pm$0.45}\\
\bottomrule[1pt]
\end{tabular}
\end{center}
\end{table}

\begin{table}[t]
\caption{Comparison to the average accuracies (\%) with 95\% confidence intervals between the methods with and without Task Aug on \textbf{miniImageNet 5-way 5-shot}.}
\label{miniImageNet_5shot}
\begin{center}
\begin{tabular}{lcc}
\toprule[1pt]
\textbf{Method} & Baseline & Task Aug \\
\hline
ProtoNets~\cite{snell2017prototypical} & 75.24$\pm$0.37 & \textbf{77.00$\pm$0.36}\\
ProtoNets (+ens) & 78.11$\pm$0.34 & \textbf{79.77$\pm$0.34}\\
ProtoNets (+val) & 76.98$\pm$0.36 & \textbf{77.59$\pm$0.37}\\
ProtoNets (+ens+val) & 79.54$\pm$0.35 & \textbf{80.60$\pm$0.34}\\
\hline
M-SVM~\cite{lee2019meta} & 77.85$\pm$0.34 & \textbf{78.90$\pm$0.34}\\
M-SVM (+ens) & 80.18$\pm$0.32 & \textbf{81.35$\pm$0.32}\\
M-SVM (+val) & 78.65$\pm$0.34 & \textbf{79.97$\pm$0.33}\\
M-SVM (+ens+val) & 81.39$\pm$0.32 & \textbf{82.13$\pm$0.31}\\
\hline
R2-D2~\cite{bertinetto2018meta} & 77.44$\pm$0.34 & \textbf{78.81$\pm$0.34}\\
R2-D2 (+ens) & 79.90$\pm$0.33 & \textbf{81.08$\pm$0.32}\\
R2-D2 (+val) & 78.61$\pm$0.35 & \textbf{79.58$\pm$0.33}\\
R2-D2 (+ens+val) & 81.34$\pm$0.32 & \textbf{81.96$\pm$0.32}\\
\bottomrule[1pt]
\end{tabular}
\end{center}
\end{table}

\textbf{Results.} Table~\ref{CIFAR-FS_1shot}-\ref{miniImageNet_5shot} show the average accuracies (\%) with 95\% confidence intervals of the methods with and without Task Aug, and the best results are highlighted. The tables show that the proposed method can improve the performance in most of cases.

We can observe that: some results without the ensemble approach~\cite{huang2017snapshot} of baseline and Task Aug are close, but the advantage of Task Aug is still obvious on the comparison results with the ensemble approach. We suspect that the scale of backbone limits the performance of the best model. A larger scale backbone is needed for the training process with Task Aug. For the results of ensemble approach, since Task Aug reduces the over-fitting, more models during the training process have good performance, which provide ensemble with models of higher quality.

Last we compare the results of this work with the results proposed by the prior works, in order to show how much this work raises the baselines after combining several prior methods and the proposed method, and they are showed in Table~\ref{CIFAR-FS_soa}, \ref{FC100_soa} and \ref{mini_soa}. The tables show that the highest accuracies of our experiments exceeded the current state-of-art accuracies 2\% to 5\%.

\begin{table}[t]
\caption{The average accuracies (\%) with 95\% confidence intervals on CIFAR-FS. $^*$CIFAR-FS results from \cite{bertinetto2018meta}. $^\dagger$Result from \cite{lee2019meta}.}
\label{CIFAR-FS_soa}
\begin{center}
\begin{tabular}{lcccc}
\toprule[1pt]
\textbf{Method} & \textbf{1-shot} & \textbf{5-shot} \\
\hline
MAML$^*$~\cite{finn2017model} & 58.9$\pm$1.9 & 71.5$\pm$1.0\\
R2-D2~\cite{bertinetto2018meta} & 65.3$\pm$0.2 & 79.4$\pm$0.1\\
ProtoNets$^\dagger$~\cite{snell2017prototypical} & 72.2$\pm$0.7 & 83.5$\pm$0.5\\
M-SVM~\cite{lee2019meta} & 72.8$\pm$0.7 & 85.0$\pm$0.5\\
\hline
M-SVM (best) (our) & 76.75$\pm$0.46 & \textbf{88.38$\pm$0.33}\\
R2-D2 (best) (our) & \textbf{77.66$\pm$0.46} & 88.33$\pm$0.33\\
\bottomrule[1pt]
\end{tabular}
\end{center}
\end{table}

\begin{table}[t]
\caption{The average accuracies (\%) with 95\% confidence intervals on FC100. $^\dagger$FC100 result from \cite{lee2019meta}.}
\label{FC100_soa}
\begin{center}
\begin{tabular}{lcccc}
\toprule[1pt]
\textbf{Method} & \textbf{1-shot} & \textbf{5-shot} \\
\hline
TADAM~\cite{oreshkin2018tadam} & 40.1$\pm$0.4 & 56.1$\pm$0.4\\
ProtoNets$^\dagger$~\cite{snell2017prototypical} & 37.5$\pm$0.6 & 52.5$\pm$0.6\\
MTL~\cite{sun2019meta} & 45.1$\pm$1.8 & 57.6$\pm$0.9\\
M-SVM~\cite{lee2019meta} & 47.2$\pm$0.6 & 62.5$\pm$0.6\\
\hline
M-SVM (best) (our) & 49.77$\pm$0.45 & 67.17$\pm$0.41\\
R2-D2 (best) (our) & \textbf{51.35$\pm$0.46} & \textbf{67.66$\pm$0.42}\\
\bottomrule[1pt]
\end{tabular}
\end{center}
\end{table}

\begin{table}[t]
\caption{The average accuracies (\%) with 95\% confidence intervals on miniImageNet. $^*$Result from \cite{lee2019meta}. Here only list the best results of previous works due to the shortage of space.}
\label{mini_soa}
\begin{center}
\begin{tabular}{lcc}
\toprule[1pt]
\textbf{Method} & \textbf{1-shot} & \textbf{5-shot}\\
\hline
\cite{gidaris2018dynamic} & 56.20$\pm$0.86 & 73.00$\pm$0.64 \\
TADAM~\cite{oreshkin2018tadam} & 58.50$\pm$0.30 & 76.70$\pm$0.30\\
LEO~\cite{rusu2018meta} & 61.76$\pm$0.08 & 77.59$\pm$0.12\\
ProtoNets$^*$~\cite{snell2017prototypical} & 59.25$\pm$0.64 & 75.60$\pm$0.48\\
M-SVM~\cite{lee2019meta} & 64.09$\pm$0.62 & 80.00$\pm$0.45\\
\hline
%M-SVM (+ens) (our) & 64.56$\pm$0.45 & 81.35$\pm$0.32\\
M-SVM (best) (our) & 65.38$\pm$0.45 & \textbf{82.13$\pm$0.31}\\
%R2-D2 (+ens) (our) & 64.79$\pm$0.45 &81.08$\pm$0.32\\
R2-D2 (best) (our) & \textbf{65.95$\pm$0.45} & 81.96$\pm$0.32\\
\bottomrule[1pt]
\end{tabular}
\end{center}
\end{table}

%\begin{table}[t]
%\caption{Comparison to the average accuracies (\%) with 95\% confidence intervals between the methods with and without Task Aug on tieredImageNet 5-way. The results are better than its compared results are highlighted.}
%\label{tieredImageNet}
%\begin{center}
%\begin{tabular}{lcccc}
%\toprule[1pt]
%\multirow{2}{*}{\textbf{Method}} & \multicolumn{2}{c}{\textbf{1-shot}} & \multicolumn{2}{c}{\textbf{5-shot}} \\
%     & Baseline & Task Aug & Baseline & Task Aug \\
%\hline
%ProtoNets~\cite{snell2017prototypical} & 62.68$\pm$0.53 & \textbf{$\pm$} & 79.64$\pm$0.40 & \textbf{$\pm$}\\
%ProtoNets (+ens) & 63.27$\pm$0.53 & \textbf{$\pm$} & 80.36$\pm$0.38 & \textbf{$\pm$}\\
%ProtoNets (+val) & 62.75$\pm$0.53 & \textbf{$\pm$} & 80.60$\pm$0.39 & \textbf{$\pm$}\\
%ProtoNets (+ens+val) & 63.44$\pm$0.53 & \textbf{$\pm$} & 80.90$\pm$0.38 & \textbf{$\pm$}\\
%\hline
%M-SVM~\cite{lee2019meta} & $\pm$ & \textbf{$\pm$} & $\pm$ & \textbf{$\pm$}\\
%M-SVM (+ens) & $\pm$ & \textbf{$\pm$} & $\pm$ & \textbf{$\pm$}\\
%M-SVM (+val) & $\pm$ & \textbf{$\pm$} & $\pm$ & \textbf{$\pm$}\\
%M-SVM (+ens+val) & $\pm$ & \textbf{$\pm$} & $\pm$ & \textbf{$\pm$}\\
%\hline
%R2-D2~\cite{bertinetto2018meta} & \textbf{65.96$\pm$0.50} &65.65$\pm$0.49 & 81.39$\pm$0.37 & \textbf{81.41$\pm$0.37}\\
%R2-D2 (+ens) & \textbf{66.18$\pm$0.50} &65.61$\pm$0.49 & \textbf{81.55$\pm$0.37} & 81.25$\pm$0.37\\
%R2-D2 (+val) & 66.54$\pm$0.50 & \textbf{$\pm$} & 81.73$\pm$0.37 & \textbf{$\pm$}\\
%R2-D2 (+ens+val) & 66.27$\pm$0.50 & \textbf{$\pm$} & 81.59$\pm$0.37 & \textbf{$\pm$}\\
%\bottomrule[1pt]
%\end{tabular}
%\end{center}
%\end{table}

\section{Conclusion}
We proposed a Task Level Data Augmentation (Task Aug), a data augmentation technique that increased the number of training classes to provide more diverse few-show task instances for meta-learning. We proved that Task Aug was valid for CIFAR-FS, FC100, and miniImageNet, and exceeded the result of the previous works. Task Aug achieved the performance by rotating the images 90, 180 and 270 degrees. This method is simple and cost-effective. With the ensemble method, we exceeded the state-of-the-art result over a large margin.

Future work will focus on searching different network structures for meta-learning, since the training with Task Aug would require larger model. Besides, we will try to apply Task Aug to other few-shot learning tasks to verify its effectiveness. Another interesting topic is to build other approaches for Task Aug, such as swapping channel order, picture blend or even auto augmentation.

{\small
\bibliographystyle{ieee_fullname}
\bibliography{paper}
}

\end{document}